\documentclass[letterpaper, 10 pt, conference]{ieeeconf}  

\IEEEoverridecommandlockouts                              

\usepackage{graphics} 
\usepackage{epsfig} 
\usepackage{mathptmx} 
\usepackage{times} 
\usepackage{amsmath} 
\usepackage{amssymb}  
\usepackage{float}
\usepackage[dvipsnames]{xcolor}
\usepackage[ruled,linesnumbered]{algorithm2e}
\usepackage{soul}
\usepackage{caption}
\usepackage{subcaption}
\usepackage{graphicx}
\usepackage{cite}

\usepackage[subtle,tracking=normal]{savetrees} 
\usepackage{relsize}
\usepackage{todonotes}
\usepackage{mathtools}
\usepackage{bm}

\DeclareMathOperator*{\argmax}{arg\,max}

\usepackage{caption}
\usepackage{subcaption}
\title{\LARGE \bf
Online Optimization of Central Pattern Generators \\ for Quadruped Locomotion
}

\author{Zewei Zhang, Guillaume Bellegarda, Milad Shafiee, Auke Ijspeert
\thanks{
This research is supported by the Swiss National Science Foundation (SNSF) as part of project No.197237. 
The authors are with the BioRobotics Laboratory, Ecole Polytechnique Federale de Lausanne (EPFL).
 {\tt \{firstname.lastname\}@epfl.ch}}
}

\begin{document}
\bstctlcite{MyBSTcontrol}

\maketitle
\thispagestyle{empty}
\pagestyle{empty}

\begin{abstract}
Typical legged locomotion controllers are designed or trained offline. This is in contrast to many animals, which are able to locomote at birth, and rapidly improve their locomotion skills with few real-world interactions. Such motor control is possible through oscillatory neural networks located in the spinal cord of vertebrates, known as Central Pattern Generators (CPGs). Models of the CPG have been widely used to generate locomotion skills in robotics, but can require extensive hand-tuning or offline optimization of inter-connected parameters with genetic algorithms. In this paper, we present a framework for the \textit{online} optimization of the CPG parameters through Bayesian Optimization. We show that our framework can rapidly optimize and adapt to varying velocity commands  and changes in the terrain, for example to varying coefficients of friction, terrain slope angles, and added mass payloads placed on the robot. We study the effects of sensory feedback on the CPG, and find that both force feedback in the phase equations, as well as posture control (Virtual Model Control) are both beneficial for robot stability and energy efficiency. In hardware experiments on the Unitree Go1, we show rapid optimization (in under 3 minutes) and adaptation of energy-efficient gaits to varying target velocities in a variety of scenarios: varying coefficients of friction, added payloads up to 15 kg, and variable slopes up to 10 degrees. See demo at: https://youtu.be/4qq5leCI2AI
\end{abstract}

\section{Introduction}
Animals display impressive abilities to swiftly refine their locomotion capabilities. Notably, newborn animals like baby goats or giraffes exhibit an innate ability to walk almost immediately post-birth, a phenomenon largely attributed to the role of Central Pattern Generators (CPGs)~\cite{doi:10.1126/science.1210617,garwicz2009unifying}. These neural networks, embedded within the spinal cord, facilitate rhythmic movements, forming the biological foundation of their rapid locomotor improvements~\cite{10.1093/acprof:oso/9780198524052.001.0001}. 

Biology-inspired control based on modeling the CPG has been extensively utilized to facilitate various types of robot locomotion. Such approaches have proven effective in controlling a wide range of systems~\cite{doi:10.1126/science.1138353,4459741, 10.1371/journal.pcbi.0030134}.
CPG-based control is characterized by stable limit cycle behavior, a limited number of control parameters, and simplicity in the integration of sensory feedback\cite{IJSPEERT2008642}. To optimize the control performance, it is essential to optimize for the best CPG parameters.  
This is usually achieved through manual tuning, evolutionary algorithms (such as genetic algorithms and particle swarm optimization)~\cite{Ijspeert1999247, Ijspeert2001331, 1249194, 973363}, and more recently with deep reinforcement learning~\cite{bellegarda2022cpgrl,shafiee2023puppeteer,bellegarda2024visual,shafiee2024viability,shafiee2024manyquadrupeds}.

\begin{figure}[t!]
    \centering
    \includegraphics[width=0.9\linewidth]{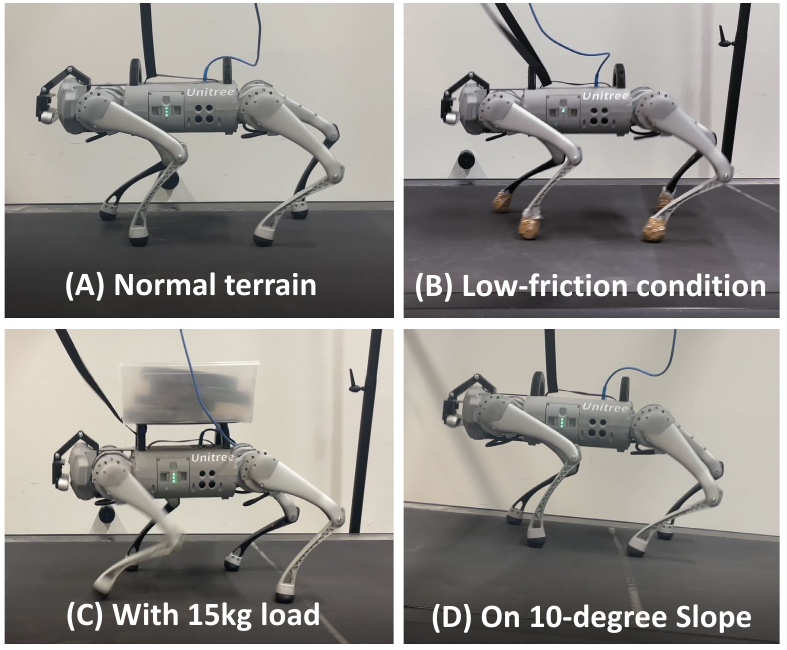}
    \caption{
    Optimized CPG-based locomotion tracking of $v_x^*$=0.4 m/s, rapidly adapting to various conditions, including normal flat terrain, slippery terrain (i.e.~taped feet), a 15 kg payload, and a 10-degree slope. 
    }
    \label{fig:cover}
    \vspace*{-1.5em}
\end{figure}

A recent surge in the utilization of Bayesian Optimization (BO) has opened new avenues for online robotic controller tuning. Through efficient interaction with the environment, BO is capable of tuning parameters at an exceptionally rapid pace. For example, Calandra et al.~demonstrate the sample efficiency of BO compared to other methods on bipedal locomotion tasks \cite{calandra2014experimental,Calandra2016}. Concurrently, Rai et al.~focus on integrating simulation data into BO to enhance sample efficiency while learning bipedal locomotion in real-world settings \cite{rai2018bayesian,rai2019using}. Moreover, Widmer et al.~leverage Contextual BO for fine-tuning MPC gains and gait parameters for trot and crawl gaits, all while adhering to safety constraints \cite{widmer2023tuning}. Our previous work applies BO for optimizing force profiles for online optimization of quadruped jumping~\cite{bellegarda2024quadruped}. 

Despite the existing success with BO, it is widely acknowledged that BO has better efficacy and performance when limiting the number of parameters that need to be tuned. Consequently, it is reasonable to anticipate good performance from using BO to tune a CPG-based controller, which has a small number of parameters~\cite{IJSPEERT2008642}. In line with this, Ruppert and Sproewitz successfully matched the CPG parameters and passive compliant dynamics, thereby enhancing locomotion energy efficiency within one hour in simulation~\cite{Ruppert2022}. 

 Building on these developments, we aim to explore the benefits of BO in the context of rapidly optimizing quadruped locomotion directly on hardware in a variety of scenarios that necessitate adaptation. In contrast to time-intensive optimization methods, we utilize BO to learn locomotion online using CPG-based control in both simulation and real-world environments, all within 
 a few minutes of physical interaction. Furthermore, we utilize Contextual Bayesian Optimization for challenging tasks to continuously learn new parameters in changing environments. Our main contributions include:

\begin{itemize}
    \item Rapid online learning of CPG parameters to locomote at varying desired velocities across terrains with varying friction,  
    and swift adaptation to constantly changing target velocities.
    \item Continuous adaptation to changing slopes and added payloads, 
    incorporating corresponding contextual information through Bayesian Optimization. 
    \item A comparison of performance across various CPG controllers, including those with and without force feedback, and with and without Virtual Model Control (VMC), in the context of locomotion optimization, showing the benefits of both force feedback and VMC for energy-efficient and stable gaits. 
\end{itemize}

The rest of this paper is organized as follows. Section \ref{sec:method} reviews background on abstract oscillators (to represent Central Pattern Generators), Virtual Model Control,  Bayesian Optimization, and Contextual Bayesian Optimization. 
Section \ref{sec:learning} presents the comprehensive optimization framework including implementation details. Section \ref{sec:results} evaluates the performance of online adaptation under various conditions in both simulation and hardware experiments, and presents the comparison results of different CPG controllers. Finally, a brief conclusion is given in Section~\ref{sec:conclusion}.
 
\section{Background} \label{sec:method}  
\subsection{Central Pattern Generators} \label{sec:CPG}

Abstract oscillators as Central Pattern Generators (CPG) are a bio-inspired approach used to generate oscillatory signals. With the coupling of abstract oscillators, synchronized behavior with a locked phase difference can produce different gaits. For quadruped robots, each leg is often represented with a single oscillator, which is coupled with other legs/oscillators. 

\subsubsection{Open-Loop CPG} 
We first consider an open-loop CPG capable of generating rhythmic motions without relying on external feedback, previously used in~\cite{bellegarda2022cpgrl}. This model has four coupled oscillators, where the equations for the amplitude and phase of oscillator $i$ are:
\begin{align}
    \dot{r}_i &= \alpha (\mu^2 - r_i^2) r_i \\
    \dot{\theta}_i &= \omega_i + \sum_{j=0}^{3} r_j w_{ij} \sin{(\theta_j - \theta_i - \phi_{ij})},
    \label{eq:coupledphase}
\end{align}
\noindent where $r_i$ is the amplitude, $\theta_i$ is the phase, $\mu$ and $\omega_i$ are the intrinsic amplitude and frequency. $\alpha$ is the convergence rate, $w_{ij}$ is the coupling weight between oscillators, and $\phi_{ij}$ is the phase lag between oscillators $i$ and $j$. The values of the intrinsic frequencies $\omega_i$ are set based on the leg phase (i.e. swing or stance phase), and we use a trot gait coupling matrix $\Phi$ for the phase lags:
\vspace{-0.5em}
\begin{equation}
    \omega_i = 
    \begin{cases}
    \omega_{swing} & \text{if } \sin{\theta_i} > 0 \\
    \omega_{stance} & \text{if } \sin{\theta_i} \leq 0
    \end{cases}; \quad
    \Phi = 
    \begin{bmatrix}
    0 & \pi & \pi & 0 \\
    -\pi & 0 & 0 & -\pi \\
    -\pi & 0 & 0 & -\pi \\
    0 & \pi & \pi & 0 \\
    \end{bmatrix}
    \label{eq:phasedot}
    \vspace*{-0.3em}
\end{equation}
where the leg order is: Front Right, Front Left, Rear Right, Rear Left. 
The oscillator states are mapped to foot trajectories (and then mapped to joint commands with inverse kinematics) with the following equations:
\vspace*{-0.4em}
\begin{align}
    x_{i, foot} &= x_{i, offset} - d_{step} r_i \cos{\theta_i}, \\
    z_{i, foot} &=
    \begin{cases}
        -h + g_c \sin{\theta_i} & \text{if $\sin{\theta_i} > 0$} \\
        -h + g_p \sin{\theta_i} & \text{otherwise}, \\
    \end{cases}
\end{align}
where $d_{step}, h, g_c, g_p$ are the desired foot step length, robot height, ground clearance, and ground penetration. The offset distance of the foot trajectory in the \textit{x} direction is denoted as $x_{i, offset}$, and varies depending on the foot index $i$, with the front legs ($i = {1, 2}$), and hind legs ($i = {3, 4}$).  

\subsubsection{Force feedback (Tegotae) in CPG}
Physical interaction can play a significant role in inter-limb coordination during leg movements. 
For example, decoupled oscillators with local force feedback can produce variable quadrupedal gaits at varying speeds~\cite{owaki2013simple}. 
To improve locomotion stability, we incorporate local force feedback into Equation \ref{eq:coupledphase} as follows:
\begin{equation}
    \dot{\theta}_i = \omega_i + \sum_{j=0}^3 r_j w_{ij} \sin{(\theta_j - \theta_i - \phi_{ij})}  -\sigma_{N} N_i \cos{\theta_i}
\end{equation}
\noindent where $N_i$ is the normal force at foot $i$, and $\sigma_N$ represents the coefficient of force feedback. Such feedback slows down the frequency when a limb is loaded, and this makes sure that a limb that carries the load of the robot (i.e.~is in stance) is not lifted prematurely (it ``waits'' until other limbs are taking over the load before switching to swing). 
As suggested in~\cite{owaki2017quadruped,thandiackal2021emergence}, we anticipate that both oscillator couplings and sensory feedback will enhance gait synchronization and robustness.

\subsection{Virtual Model Control} 

Virtual Model Control (VMC) has previously been used for enhancing the capabilities of quadruped robots, especially for stability on challenging terrains \cite{VMC1,VMC2}. It is a simple and effective approach for posture control. Augmenting the CPG with VMC helps to correct the contact forces based on the current robot attitude and heading. 
We use the attitude and direction controllers introduced in \cite{VMC1} to control the orientation (roll, pitch, yaw) of the trunk. 

\subsection{Bayesian Optimization} \label{sec:BO}

Bayesian optimization (BO) is a sequential model-based optimization approach used to find optimal parameters through fitting the unknown model with sampling~\cite{frazier2018tutorial}. This sample-efficient optimization approach can compute the delegate model based on all input-output pairs obtained from strategically sampling the environment. The general form of the Bayesian Optimization problem is written as: $\mathbf{x} = \argmax_{\mathbf{x}} \hat{f}(\mathbf{x})$, 
where $\mathbf{x}$ denotes the parameters to be optimized, and $\hat{f}(\cdot)$ represents the surrogate model of the unknown function based on the sampling.

Several methods can be used for acquiring such a delegate model, for example Gaussian Processes (GP) and Tree Parzen Estimators (TPE). Maximizing the acquisition function determines which set of parameters $\mathbf{x}$ to observe next. There are many options for the acquisition function such as Upper Confidence Bound (UCB), Expected Improvement (EI), and Probability Improvement (PI). In this work, we use Gaussian Processes to regress all of the samples, and we select UCB as our acquisition function defined as: $a(\mathbf{x};\beta) = \mu(\mathbf{x}) + \beta \sigma(\mathbf{x})$. 
Here, $\mu(\mathbf{x})$, $\sigma(\mathbf{x})$ denote the mean and standard deviation of the Gaussian Process, respectively. $\beta$ is an explicit parameter designed to balance exploitation and exploration in UCB, and we define it as a decaying parameter. Further details regarding $\beta$ will be elaborated in Section \ref{sec:opt_setting}.

\subsection{Contextual Bayesian Optimization} \label{sec:CBO}
Contextual Bayesian Optimization (CBO) extends standard Bayesian Optimization (BO) to address multi-task problems occurring in different contexts, as opposed to single-task scenarios \cite{multitask}. By incorporating additional contextual variables, which typically relate to environmental information, CBO enables the derivation of specific solutions tailored to corresponding environments. 
CBO and the corresponding UCB acquistion function can be determined by modifying the standard BO formulation:
\begin{equation}
\vspace*{-0.4em}
    \mathbf{x} = \argmax_{\mathbf{x}} \  \hat{f}(\mathbf{x}, \mathbf{c}),
    \label{eq:CBO}
\end{equation}
\begin{equation}
    a(\mathbf{x};\beta, \mathbf{c}) = \mu(\mathbf{x; \mathbf{c}}) + \beta \sigma(\mathbf{x; \mathbf{c}})
    \label{eq:UCB_CBO},
\vspace*{-0.3em}
\end{equation}

\noindent where $\mathbf{c}$ denotes the contextual variable that relates to the general environmental conditions. This approach also enhances sample efficiency by leveraging optimization results from nearby contexts~\cite{DBLP:journals/corr/abs-1803-00196}. Our framework is grounded in the theoretical framework proposed by Krause and Ong~\cite{CBOKrause}, which involves constructing a product of kernels of the optimized and contextual parameters.

\section{Optimization Framework} 
\label{sec:learning}
\begin{figure}
    \centering
  \includegraphics[clip, trim=0cm 0cm 0cm 0cm, width=0.95\linewidth]{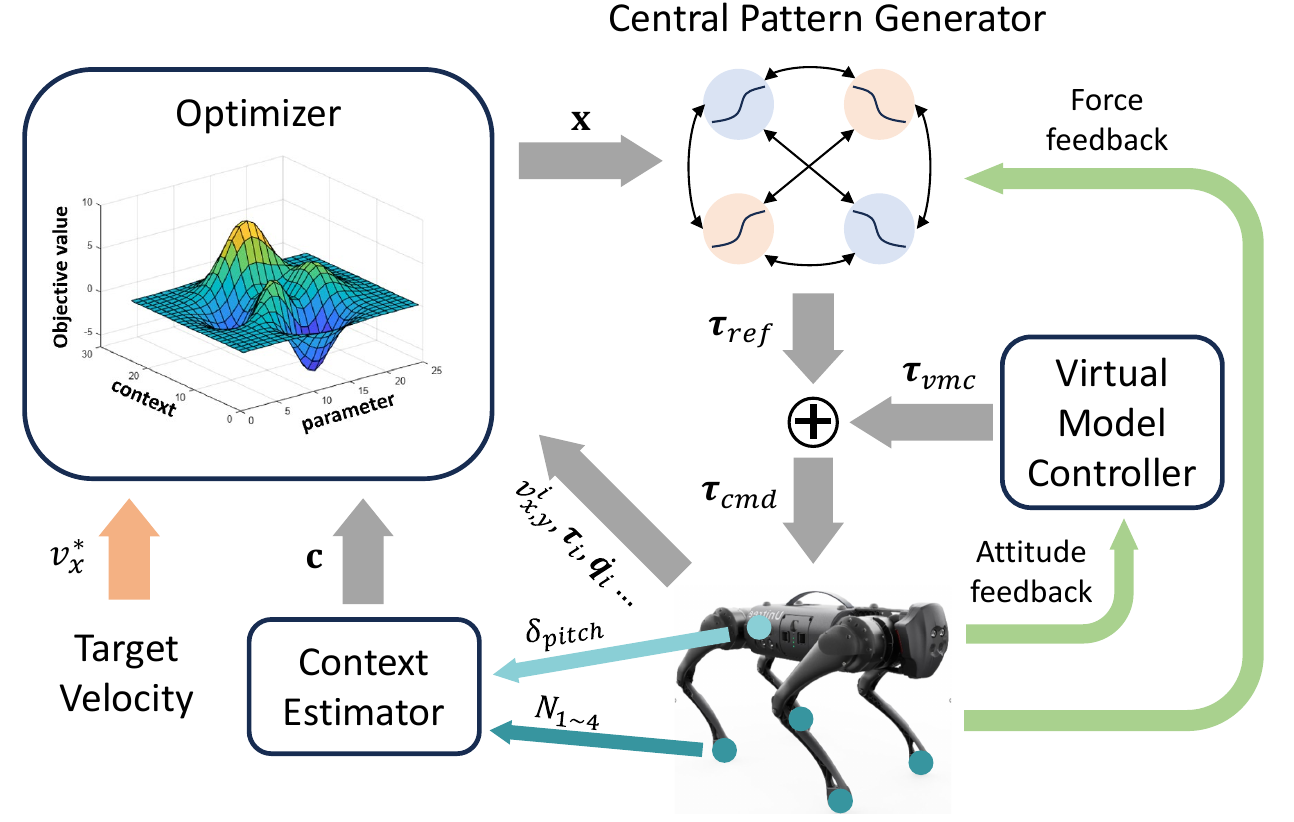}\\
  \vspace{-0.5em}
  \caption{Control architecture for online CPG optimization. The optimizer updates the CPG parameters $\mathbf{x}$ at the beginning of each trial based on the current target velocity $v_x^*$ and contextual information $\mathbf{c}$. It receives the base velocity and joint data during the steady-state phase of each trial to compute the objective value. Force feedback to the CPG phase equations and Virtual Model Control help to stabilize the robot and promote energy-efficient gaits. 
  }
\label{fig:architecture}
\vspace*{-1.5em}
\end{figure}

In this section we detail our Contextual Bayesian Optimization framework to tune our CPG-based controller for energy-efficient quadruped locomotion to track varying target velocities in a variety of scenarios.
A high-level control diagram is depicted in Figure \ref{fig:architecture}, and we explain all components below. 
\subsection{Locomotion Optimization}
In each optimization loop, the optimizer selects a new set of CPG parameters based on the velocity command input from the user, estimated contextual information, and the objective value computed from data collected in the previous cycle. 
We optimize eight parameters introduced in Section \ref{sec:CPG} during the optimization: $\mathbf{x} = [g_c, g_p, \omega_{swing}, \omega_{stance}, \mu, x_{offset,front}, x_{offset,hind}, \sigma_N]$. The ranges $\mathcal{X}$ for these parameters are specified in the Appendix in Table \ref{tab:params_range}, and additional fixed control variables are detailed in Table \ref{tab:control_params}. 
With the updated parameters, the CPG module cyclically outputs reference joint torques $\bm{\tau}_{ref}$, computed with inverse kinematics and PD control based on the reference foot trajectory of the CPG with force feedback. The final command torques, $\bm{\tau}_{cmd}$, is the sum of the reference torques from the CPG $\bm{\tau}_{ref}$ and the VMC torques $\bm{\tau}_{vmc}$, which are input to the robot for the locomotion task.

\subsubsection{Objective Function}
We define a simple objective function, and the optimal parameters $\mathbf{x}^*$ will be found to maximize the objective value computed as follows:
\begin{equation}
\small
     J = w_1 \Delta t \mathlarger{\mathlarger{\sum}}_{i=0}^T \min \left (\exp{\left(-\frac{\lVert v^{i}_{x,y} - v^*_{x,y} \rVert ^ 2}{0.05}\right)}, l_r \right) - w_2 \mathrm{CoT},
    \label{eq:costfunction}
\vspace*{-0.3em}
\end{equation}

\noindent where the weights $w_1, w_2$ are equal to 1 and 0.5 respectively, $v^i_{x,y}$ is the instantaneous body linear velocity, $\Delta t$ is the time step (1 ms), $T(=3000)$ is the total number of steps in one trial, $l_r$ is the maximum velocity reward. The function consists of two components. The first component is the velocity reward, which increases as the actual velocity comes closer to the target one $v_{x,y}^*$ (=$[v_x^*, 0]$). We include the lateral velocity $v_y$ in the function to penalize lateral movement, since some sets of parameters could lead to unexpected oscillations or drift in the lateral direction. The second component aims to lower the energy consumption by penalizing the Cost of Transport (CoT)\cite{gabrielli1950price} in each trial, defined as follows:
\begin{equation}
    \mathrm{CoT} = \frac{\sum_{i=0}^T \langle \lvert \bm{\tau}_{i} \rvert, \lvert \bm{\dot {q}}_{i} \rvert \rangle \Delta t}{mgd},
    \label{eq:CoT}
\end{equation}
\noindent where $\bm{\tau}_i$ and $\bm{\dot q}_i$ denote the vectors of the torque and joint velocity at time $i$, 
$m$ is the total mass of the robot and any payload, $g$ is the gravitational constant, and $d$ is the distance covered in one trial. 
The maximum velocity reward is bounded at $l_r=0.85$ to expand the range of acceptable actual velocities around the target. This adjustment enables the optimizer to concentrate more on reducing energy consumption when the velocity is close to the target. Through the optimization of this function, we aim to fine-tune the parameters such that the robot achieves locomotion at the target velocity while minimizing energy consumption. 

\subsubsection{Data Reuse} \label{sec:data_reuse}
We implement a data reuse technique to enhance the sample efficiency of our optimizer when transitioning between different target velocities. Upon receiving a new target velocity from the user, we leverage the data previously collected during the optimization for the former target. This approach allows the optimizer to recalibrate all prior objective values by merely adjusting the desired velocity value in Equation \ref{eq:costfunction}. Such re-utilization of existing data facilitates rapid adaptation to the new velocity, eliminating the necessity to restart the optimization process with each change in velocity target.

\subsubsection{Context Estimator}
As previously mentioned, CBO uses contextual variables to achieve multi-task learning. Therefore, we propose a context estimator to estimate contextual information from the environment. We take into account two contextual variables $\mathbf{c}=[c_{load}, c_{slope}]$ which relate to the load/mass of the robot and the slope information of the terrain. We implement two simple estimators to measure this information during each optimization cycle: 
\begin{equation}
    c_{load} = \frac{1}{4T} \sum_{i=0}^T \sum_{j=0}^4 N_{i,j} \ , \quad
    c_{slope} = \frac{1}{T} \sum_{i=0}^T \delta_{pitch, i}
    \label{eq:context_est}
\end{equation}
where $c_{load}$ estimates the average normal force measurement $N_{i,j}$ collected by the force sensors on all four feet, while $c_{slope}$ estimates the average pitch measurements $\delta_{pitch,i}$ obtained from the IMU. Both estimators are used to provide a two-dimensional contextual variable $\mathbf{c}$ to the optimizer throughout the experiment. As demonstrated in Section \ref{sec:results}, each estimator is capable of providing consistent contextual information during the optimization process,
which is crucial for learning
the optimal parameters under varying contexts.
\RestyleAlgo{ruled}
\SetKwComment{Comment}{/* }{ */}
\setlength{\textfloatsep}{0pt}
\begin{algorithm}[t!]
\small
\caption{\small 
Online optimization for quadruped locomotion learning with CPG using CBO. 
}\label{alg:CBO_loco}
\textbf{Input: }$\mathbf{x}_{init}\in \mathcal{X}$, $\beta_{init}$, $N$, $v_x^*$\;
Initialize datasaver of objective values and parameters $\mathbb{S} \gets \emptyset$,  $\mathbb{P} \gets \emptyset$\;
$i \gets 0$; $\mathbf{x} \gets \mathbf{x}_{init}$; $\beta \gets \beta_{init}$\;
\While{$i \leq N$}{
\If{$v_x^*$ is updated}{$\mathbb{S} \gets $ Update previous objective values with $v_x^*$\;}
$\mathbf{c}$,  $J_i \gets$ Update current context and objective value\;
$\mathbb{S} \gets \mathbb{S} \cup J_i$; $\mathbb{P} \gets \mathbb{P} \cup [\mathbf{x}, \mathbf{c}]$\;
$\hat{f} \gets $ Update surrogate model with $\mathbb{S}$, $\mathbb{P}$\;
$\beta \gets $ Update with Equation \ref{eq:decay_beta}\;
  \eIf{$i \leq 2 $}{
    $\mathbf{x} \gets random(\mathbf{x}\in\mathcal{X})$\; 
  }{
    $\mathbf{x} \gets \text{argmax}_{\mathbf{x}} a(\mathbf{x}; \beta, \mathbf{c})$
  }
}
\end{algorithm}
\subsection{Algorithm} 
\label{sec:opt_setting}
Algorithm \ref{alg:CBO_loco} summarizes the overall optimization procedure with CBO. For the initial two parameter updates, we randomly sample new points. Beyond this, the acquisition function plays a crucial role in updating parameters for subsequent trials. With the UCB acquisition function, we achieve a balance between exploration and exploitation by adjusting the explicit hyperparameter $\beta$, as detailed in Equation \ref{eq:UCB_CBO}. We employ a time-decaying variable $\beta$ to foster increased exploration in the early stages of optimization. This approach gradually shifts the emphasis towards exploitation by reducing $\beta$'s value, thus avoiding overly aggressive sampling. The update of the parameter $\beta$ is governed by Equation \ref{eq:decay_beta}:
\begin{equation}
    \beta = \beta_{init}\cdot \gamma^{\max(n-n_{decay}, 0)},
    \label{eq:decay_beta}
\end{equation}
\noindent where $\beta_{init}, \gamma, n_{decay}, n$ represent the initial value of $\beta$, the decay rate, the trial number at which the decay begins, and the total number of trials considered in the current context, respectively. The values of these parameters and details for updating them are presented in the Appendix (Section~\ref{app:contextsharing}). 

\vspace*{-0.1em}
\subsection{Implementation Details}
\vspace*{-0.1em}
We designate a short-length horizon for one optimization cycle. Each trial is divided into two stages: the transition phase, 
which lasts for 1.5 seconds, and the steady-state phase, lasting for 3 seconds. The computation of the objective value and contextual variables is based solely on the data collected during the steady-state phase, in order to minimize the influence of parameter transitions. 
In the transition phase, we implement a simple linear interpolation to smooth the parameter transition between two trials. 

Our optimization framework is implemented in the BoTorch framework \cite{balandat2020botorch} and is built upon the \textit{bayesopt4ros} package, developed by Fröhlich et al.~\cite{fröhlich2021model}. Compared to  their framework, we include the decaying exploration factor $\beta$ in the acquisition function, and apply the data reuse technique explained in Sec.~\ref{sec:data_reuse} to improve the sample efficiency.

\begin{figure*}[t]
    \centering
  \includegraphics[clip, trim=0cm 0cm 0cm 0cm, width=0.95\textwidth]{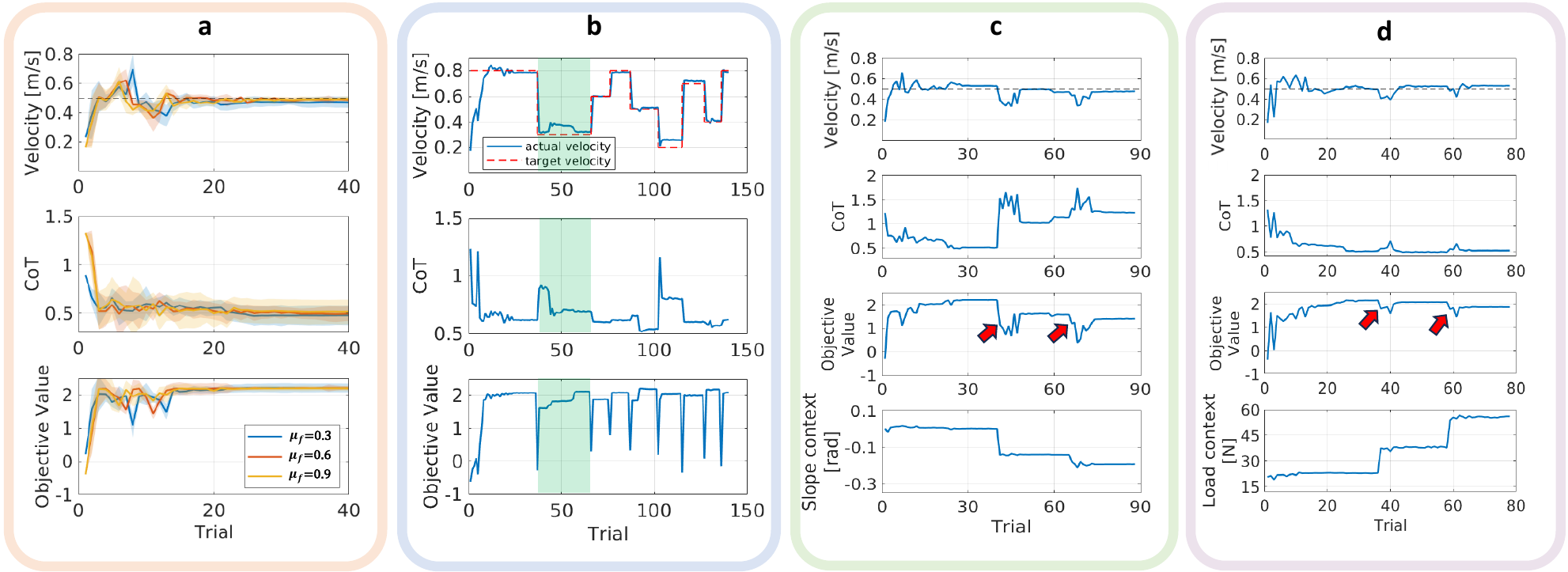}
  \vspace{-0.3em}
  \caption{
  Simulation results for online locomotion learning, where we report the mean forward velocity, Cost of Transport (CoT), objective value, and context information in different scenarios. \textbf{(a)}: adapting to different coefficients of friction while tracking a target velocity of 0.5 m/s. \textbf{(b)}: rapid adaptation to moving target velocities. \textbf{(c)}: tracking 0.5 m/s with adaptation from flat terrain to varying slopes of 10 and 12.5 degrees. \textbf{(d)}: tracking 0.5 m/s with adaptation to added loads of 7.5 kg and 15 kg. Red arrows indicate the transition time to a new contextual condition (slope or load). 
  }
\label{fig:sim_overview}
\vspace*{-2em}
\end{figure*}
\subsection{Experimental Setup}
We test our pipeline on the Unitree Go1 in both simulation and on hardware. For simulations, we use Gazebo with the ODE physics engine. In hardware experiments, we conducted the optimization trials with the robot placed on a treadmill. 
All proprioceptive sensing measurements are read by the Unitree sensors, namely the IMU, joint states, and foot contact sensors.

\section{Results} 
\label{sec:results}
In this section, we report results from rapidly optimizing locomotion tasks in both simulation and hardware experiments.  
We demonstrate our framework's ability to optimize for energy-efficient single velocity tracking tasks, as well as its ability to adapt to changes in the environment in several different scenarios, including:
adaptation to varying coefficients of friction, varying velocity commands, varying terrain slopes, and added payloads of up to 15 kg.
We also conduct a comparative analysis of the CPG controller's performance with respect to both force feedback and the attitude controller (VMC), and discuss the effectiveness of each controller's corresponding module within our framework. The reader is encouraged to watch the supplementary video for clear visualizations of the discussed experiments. 

\subsection{Learning Single Velocity Tracking}

In simulation, we define three types of terrains with friction coefficients of 0.3, 0.6, and 0.9, which can correspond to slippery, normal, and coarse terrain. We optimize the parameters of the CPG with both the force feedback and VMC controller to test the learning performance to track a target forward velocity of 0.5 m/s. 
We randomly initialize the parameters, which gives a low initial objective value. 
As depicted in Figure \ref{fig:sim_overview}a, our pipeline exhibits rapid adaptability to diverse terrains, as evidenced by the average performance obtained from five random seeds. We observe the rapid improvement of the objective value within very few trials, with 40 trials taking \textbf{less than 3 minutes}. For all cases, multiple prospective solutions (with high objective value) were identified within fewer than 10 trials (45 seconds). Ultimately, the optimizer consistently outputs a set of parameters that maximizes the objective value. 
This result indicates the system's capability to track the target velocity and improve energy-efficiency effectively. 

\subsection{Multi-Velocity Tracking}

Our framework enables the robot to rapidly adapt to different target velocities on a single terrain, leveraging data reuse as discussed in Section \ref{sec:data_reuse}. Figure \ref{fig:sim_overview}b shows the robot successfully learns to adapt its gait for consecutively changing target velocities. Initially, the robot learns to walk at a target speed of 0.8 m/s from scratch. Subsequently, when the target velocities were altered to 0.3 m/s and then to 0.6 m/s, the robot required fewer trials to achieve these speeds, benefiting from the knowledge acquired in previous learning phases. Furthermore, the CoT curve between Trials 40 and 60 (green highlight in the plot) 
illustrates the robot's capability to continuously improve gait energetic efficiency by fine-tuning around the most promising parameters. Within a learning span of approximately 140 trials (about 10 minutes), our framework demonstrates a comprehensive process through which the robot evolves from walking at a single velocity to efficiently tracking multiple velocities. 

\subsection{Slope and Load Adaptation} \label{sec:slope_load_adapt}
Our framework's adaptability to new environmental conditions, including changes in slope and added load, is further demonstrated with estimated contextual information.

Figure \ref{fig:sim_overview}c shows the robot's learning process to maintain a target velocity of 0.5 m/s while transitioning from flat terrain to inclines of 10 and 12.5 degrees.
The plot of contextual variable $c_{slope}$ confirms the reliability of our context estimator across single terrains, with $c_{slope}$ values inversely correlating with increased slope angles. Notably, upon encountering new terrain types, there is an initial drop in the objective value, followed by subsequent improvements after several optimization cycles. 

To verify the robustness of the learned parameters, as shown in the video, we then tested our optimized controller on slopes gradually increasing from 10 to 17.5 degrees in inference mode, by fixing $\beta$ in Equation \ref{eq:UCB_CBO} to $10^{-6}$ to minimize exploration risk. 
We observe that although the robot had only optimized parameters to locomote on a maximum slope of 12.5-degrees, it is still able to adapt to slopes of up to 15 degrees, highlighting our framework's capability to efficiently select learned parameters based on the estimated contextual information. 
It also underscores the potential to select viable parameters for conditions which were not seen duirng the optimization.

For the load adaptation experiment in Figure \ref{fig:sim_overview}d, we first add a 7.5 kg load (at Trial 36), followed by an additional load to reach a total of 15 kg (at Trial 58). The target velocity for this experiment was also set to 0.5 m/s. We observe the robot is able to adapt under progressively increasing payload conditions, quickly adapting the parameters to maintain the target velocity and lower energy consumption. Table \ref{tab:load_adapt} showcases the significant improvements in velocity tracking, CoT, and objective value following adaptation compared with the performance using the parameters optimized under normal conditions, thereby confirming the effectiveness of our framework.

\begin{table}
    \vspace{0.3em}
    \centering
    \caption{Mean velocity, CoT, and objective value measured under 0 kg, 7.5 kg, and 15 kg load conditions with the CPG parameters before and after adaptation with our framework. Values in parentheses represent outcomes obtained by directly applying the optimal parameters identified on flat terrain without any load for the target velocity $v_x^*=0.5$ m/s. 
    }
    \vspace*{-.3em}
    \begin{tabular}{|c|c|c|c|}
    \hline
       Load [kg]  & Velocity  [m/s] & CoT & Objective Value  \\
    \hline
        0 & 0.491 (-) & 0.510 (-) & 2.175 (-) \\
    \hline
        7.5 & 0.524 (0.403) & 0.489 (0.564) & 2.213 (1.9735) \\
    \hline
        15 & 0.528 (0.341) & 0.523 (0.905) & 2.189 (1.270) \\
    \hline
    \end{tabular}
    \label{tab:load_adapt}
    \vspace*{-.0em}
\end{table}

\begin{figure*}[!t]
    \centering
  \includegraphics[clip, trim=0cm 0cm 0cm 0cm, width=0.9\textwidth]{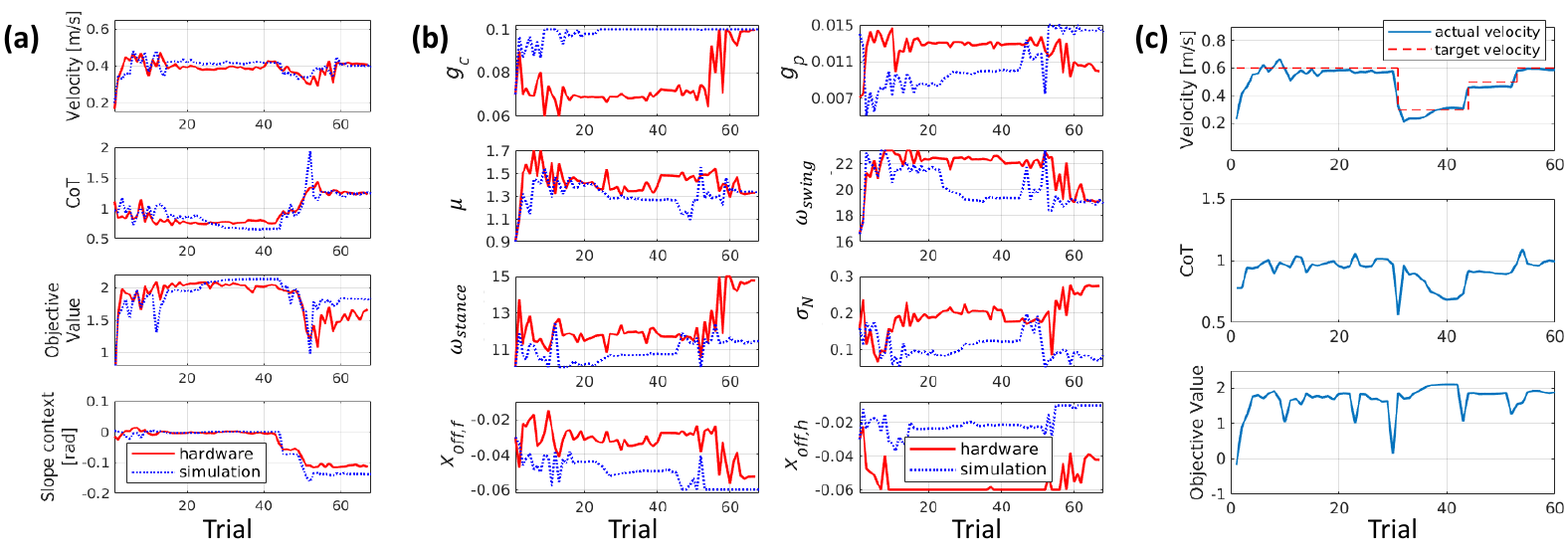}\\
  \vspace{-0.6em}
  \caption{
  Slope adaptation and velocity tracking in hardware experiments. \textbf{(a)} mean velocity, CoT, objective value, and slope context, increasing the slope to 10 degrees while tracking 0.4 m/s. 
  \textbf{(b)} optimized parameters evolution during \textbf{(a)}, from flat terrain locomotion to slope adaptation.
  \textbf{(c)} adaptation to changing velocity commands (0.6$\rightarrow$0.3$\rightarrow$ 0.5$\rightarrow$0.6 m/s) on the hardware.   
  }
\label{fig:slope_adapt_exp}
\vspace*{-1.3em}
\end{figure*}
\subsection{Hardware Experiments} \label{sec:result_real_world}

\begin{figure}[!t]
\vspace{-0.4em}
    \centering
    \includegraphics[clip, trim={0.2cm 0.0cm 0.0cm 0.0cm},width=0.85\linewidth]{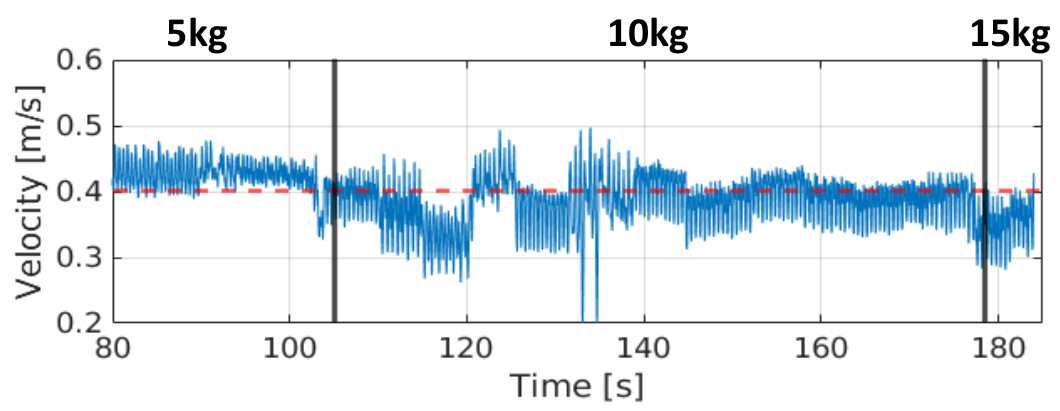} \\
    \vspace{-0.4em}
    \caption{Forward velocity during the hardware experiment of adaptation up to a 15 kg load, tracking a target velocity of 0.4 m/s. Before this adaptation, the robot was trained on flat terrain without any load for three minutes. 
    }
    \label{fig:load_exp_vel}
\end{figure}

We validate our framework's effectiveness of learning optimal locomotion and adaptation in hardware experiments using the CPG with force feedback and VMC on the Unitree Go1 robot. Our experiments successfully demonstrate:

\begin{itemize}
    \item single velocity tracking at 0.4 m/s with CoT $\approx$ 0.8 on normal terrain and with CoT $\approx$ 0.95 on slippery terrain
    \item multi-velocity tracking, transitioning from 0.6 m/s (CoT $\approx$ 0.95), to 0.3 m/s (CoT $\approx$ 0.70), to 0.5 m/s (CoT $\approx$ 0.90), Figure \ref{fig:slope_adapt_exp}c
    \item adaptation to a 10-degree slope, Figure \ref{fig:slope_adapt_exp}a and \ref{fig:slope_adapt_exp}b
    \item adaptation to increasing loads up to 15 kg, Figure~\ref{fig:load_exp_vel}
\end{itemize}

Figures \ref{fig:slope_adapt_exp}a and \ref{fig:slope_adapt_exp}b demonstrate how our framework allows the Go1 robot to adapt online to a 10-degree slope while tracking a target velocity of 0.4 m/s. 
The optimizer first converges to a set of parameters which facilitates tracking of the target velocity on flat terrain within fewer than 40 trials (\textbf{less than 3 minutes}) of learning. 
Subsequently, after the slope was increased to 10 degrees, the robot stabilizes around this target once more within approximately 20 trials (\textbf{around 2 minutes}). 
Figure \ref{fig:slope_adapt_exp}b compares the final parameters that enable reaching the target velocity on the 10-degree slope after adaptation, with those optimized for flat terrain in the hardware experiment. 
We observe a decrease in $\omega_{swing}$, $x_{offset, front}$, and $g_p$. Conversely, there is an increase in $\omega_{stance}$, $g_c$, $\sigma_N$, and $x_{offset, hind}$. When comparing the changes in parameters between the simulation and the hardware experiment, we find that $g_c, \mu, \omega_{swing}, x_{offset, front}$ converge to similar values under both conditions. Meanwhile, $x_{offset, hind}$ in the hardware experiment demonstrates a similar (but shifted) trend to that observed in simulation during adaptation.

Figure \ref{fig:load_exp_vel} shows the robot base linear velocity after placing loads of 5 kg, 10 kg, and 15 kg on the robot, which began on flat terrain without any load to track a velocity of 0.4 m/s. This result demonstrates the ability of our framework to enable the robot to adaptively locomote under various load conditions through the online optimization.

\begin{figure}[t!]
    \centering
    \vspace{-0.4em}
    \includegraphics[clip, trim=0.cm 0.0cm 0.cm 0.0cm,width=0.9\linewidth]{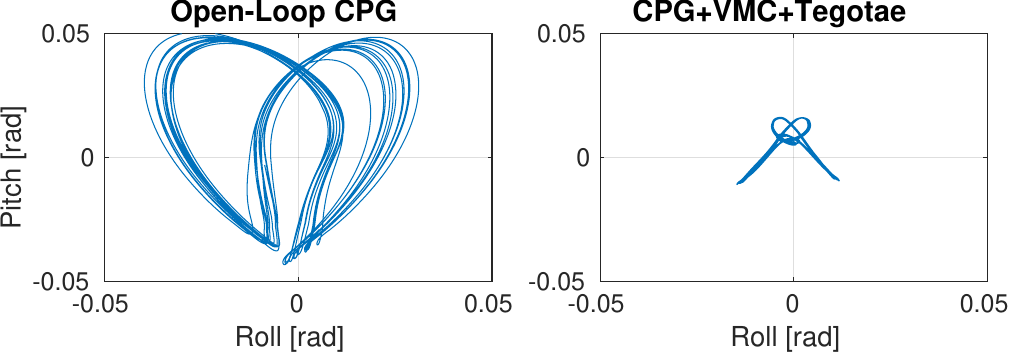}\\
    \vspace{-0.2em}
    \caption{
    Comparison of roll-pitch variations of the optimized open-loop CPG, and of the CPG with force feedback (Tegotae) and Virtual Model Control after learning in simulation. 
    }
    \label{fig:att_comp}
\vspace*{-0.em}
\end{figure}
\subsection{Ablation Study of CPG Controllers}

We perform ablation studies towards understanding the importance of augmenting the CPG with both force feedback and posture control. We repeat the optimization by optimizing parameters for the following four controllers: 1) Open-Loop CPG, 2) CPG+VMC, 3) CPG+Tegotae, and 4) CPG+VMC+Tegotae (our method). When Tegotae is not active, we optimize one less parameter, since $\sigma_N$ is not needed. 
Table \ref{tab:cont_comp} compares the final average performance of five optimization runs with different random seeds for each controller type in maintaining the same target velocity 0.5 m/s. 

As demonstrated in Table \ref{tab:cont_comp}, controllers incorporating Tegotae (force feedback) consistently achieve lower CoT with their final parameters compared to their counterparts. This outcome suggests that force feedback could contribute to generating higher-quality and more energy-efficient gaits, as reflected by the lower average final CoT. 
While the VMC component alone appears to be less critical than Tegotae force feedback in enhancing gait quality — particularly when comparing the CPG controllers with and without VMC — the results indicate that combining VMC with force feedback further improves energy efficiency.

Moreover, controllers lacking force feedback exhibited correspondingly lower objective values, and higher standard deviations of these values, indicating less consistent and more variable optimization performance. This instability could stem from a reduced area of promising parameters in the search space for controllers without force feedback, thereby complicating the identification of optimal parameters within a limited time frame. 

Figure \ref{fig:att_comp} compares the roll-pitch trajectory of the robot's attitude when applying the best individual parameters for both the Open-Loop CPG controller and the CPG+VMC+Tegotae controller after optimization. The force feedback (Tegotae) and VMC greatly improve the stability by reducing the roll and pitch variations, contributing to better energy efficiency and velocity tracking. 

\begin{table}[t!]
\vspace{0.06in}
    \caption{Mean velocity, CoT and objective value using different CPG controllers for the target velocity $v_x^*=0.5$ m/s. These results were derived by averaging the outcomes of the last five trials in a single experiment and then further averaging these across five experiments, each with distinct random seeds. Each experiment consisted of 40 trials. 
    }
    \vspace{-0.3em}
    \centering
    \begin{tabular}{|c|c|c|c|}
    \hline
       Controller  & Velocity [m/s] & CoT & Objective Value \\
    \hline
        OpenLoop & 0.477 $\pm$ 0.045 & 0.931 $\pm$ 0.138 & 1.638 $\pm$ 0.516 \\
    \hline
        VMC & 0.494 $\pm$ 0.028 & 0.919 $\pm$ 0.149 & 1.643 $\pm$ 0.139 \\
    \hline
        Tegotae & 0.500 $\pm$ 0.012 & 0.583 $\pm$ 0.024 & 2.153 $\pm$ 0.042 \\
    \hline
        VMC+Tegotae & 0.509 $\pm$ 0.023 & 0.518 $\pm$ 0.050 & 2.167 $\pm$ 0.072 \\
    \hline
    \end{tabular}
    \label{tab:cont_comp}
    \vspace*{0.5em}
\end{table}

\section{Conclusion}\label{sec:conclusion}
In this study, we introduced a framework capable of efficient online learning of CPG-based locomotion for quadruped robots, both in simulation and in hardware experiments, utilizing Bayesian optimization. This is in contrast to other approaches for optimizing the CPG parameters, typically done through manual tuning or offline optimization. Our results demonstrate that our framework facilitates rapid learning and adaptation to varying target velocities, and to terrains with varying coefficients of friction. 
Furthermore, by introducing contextual variables into Bayesian optimization, we achieve rapid adaptation to a 10-degree slope and a 15 kg load (125\% of the nominal mass) on the Unitree Go1 hardware. To the best of our knowledge, this represents the highest added payload ever achieved on the Go1 robot. Moreover, our findings suggest that energy efficiency and gait stability are much improved by incorporating force feedback (as well as adding posture control (VMC)) into the CPG. 

Current limitations of this work include the absence of explicit safety guarantees during parameter searching, as well as the potential applicability to more rugged terrain. 
Future work will aim to incorporate more advanced feedback controllers and BO with safety constraints to address these issues. We also plan to learn parameters for different gaits, as well as optimize joint-space CPGs, towards a better understanding of the biological parallels in learning locomotion. 




\bibliographystyle{IEEEtran}
\bibliography{ref}
\section{Appendix}
\subsection{Parameter Setting} \label{app:range_context}

\noindent 
\begin{minipage}{.48\linewidth}
\captionof{table}{Optimization parameter ranges} 
\centering
\resizebox{\columnwidth}{!}{
\begin{tabular}{c|c}
 \hline
    $g_c$ [m] & [0.06, 0.10]  \\
 \hline
    $g_p$ [m] & [0.005, 0.015]  \\
 \hline
    $\omega_{swing}$ [rad/s] & [16, 23]  \\
 \hline
    $\omega_{stance}$ [rad/s] & [10, 15]  \\
 \hline
    $\mu$ & [0.9, 1.7] \\
 \hline
    $x_{offset, front}$[m] & [-0.06, -0.01] \\
 \hline
    $x_{offset, hind}$[m] & [-0.06, -0.01] \\
 \hline
    $\sigma_N$ & [0.05, 0.30] \\
 \hline
\end{tabular}
}
\label{tab:params_range}
\end{minipage}%
\hfill 
\begin{minipage}{.48\linewidth}
\captionof{table}{Control parameters} 
\centering
\resizebox{.55\columnwidth}{!}{
\begin{tabular}{c|c}
 \hline
 $w_{ij}$ & 1 \\
 \hline
 $\alpha$ & 50 \\
 \hline
$d_{step}$ [m] & 0.05 \\ 
 \hline
$h$ [m] & 0.3\\ 
\hline
    $f$ [kHz] & 1 \\
\hline
   $k_{dir}$  & 300 \\
 \hline
    $k_{att,r}$ & 150 \\
 \hline
    $k_{att,p}$ & 160 \\
 \hline
    $K_{p, joint}$ & 70 \\
 \hline
    $K_{d, joint}$ & 1.3 \\
\hline
\end{tabular}
}
\label{tab:control_params}
\vspace*{+1em}
\end{minipage}

\subsubsection{Optimization Parameter Ranges}
The ranges used for the optimization parameters in both simulation and hardware experiments are provided in Table \ref{tab:params_range}.

\subsubsection{Control Parameters}
In Table \ref{tab:control_params}, $f$ represents the control frequency, and $k_{dir}, k_{att,r}, k_{att,p}$ correspond to the gains for yaw, roll, and pitch control for the VMC controller, respectively. 

\subsubsection{Parameter Normalization}
Both optimization and context parameters require normalization based on certain ranges before computing the surrogate model with a Gaussian Process. We normalize optimization parameters within the ranges provided above, and contextual variables as follows: average normal force $c_{load} \in [15, 55]$, average pitch of the trunk $c_{slope} \in [-0.4, 0.1]$. These ranges are determined based on the tasks and physical sensor capabilities of the Unitree Go1 robot, ensuring that the contextual variables remain within realistic bounds. 

\subsection{Exploration Factor Updating}
\label{app:contextsharing}
\subsubsection{Parameters used in the UCB function}
The values of these parameters can be found in Table \ref{tab:opt_param}. After starting the optimization, we use the value of $\beta_{init}$ and $n_{decay}$ in the table in Eq.~\ref{eq:UCB_CBO}. When the estimator detects changes in the contextual variables $c_{load}$ and $c_{slope}$ during the optimization, the exploration factor $\beta$ is updated based on the smaller value $\beta_{init, c}$, $n_{decay, c}$, which is shown as follows: 
\begin{align}
    \beta = \beta_{init,c}\cdot \gamma^{\max(n-n_{decay,c}, 0)}
\end{align}
This is done because given the optimizer's ability to rapidly identify a suitable set of gait parameters for new contexts — leveraging knowledge from previous contexts via CBO — a reduced emphasis on exploration could be warranted. Moreover, excessive exploration, especially during such adaptations, could cause a robot fall due to the increased task difficulty. Consequently, we adopt a strategy that prioritizes reduced exploration during contextual adaptation, enhancing safety within our optimization framework.
\begin{table}[!t]
\setlength{\tabcolsep}{20pt}
    \caption{Parameters used to update exploration factor $\beta$ }
    \vspace{-0.1em}
    \centering
    \begin{tabular}{c|c}
     \hline
        $\beta_{init}$ & 5  \\
     \hline
        $\beta_{init, c}$  & 1.5  \\
     \hline
        $\gamma$ & 0.7  \\
     \hline
        $n_{decay}$ & 10  \\
     \hline
        $n_{decay, c}$ & 3  \\
     \hline
    \end{tabular}
    \label{tab:opt_param}
\vspace{0.4em}
\end{table}

\subsubsection{Determination of Context Sharing} 
 The assessment of whether trials occur within the same context is based on predefined thresholds, $t_{slope}$ and ${t_{load}}$. A trial is considered to be in the same context if it meets the following criterion:
\begin{equation}
\vspace*{-0.4em}
    n = \text{count}
    \begin{rcases}
    \begin{dcases}
    c_{load}^i-t_{load}r_{load} \leq \mathbf{c}_{load} \leq c_{load}^i+t_{load}r_{load} \\
    \qquad \qquad \qquad \qquad \text{and} \\
    c_{slope}^i-t_{slope}r_{slope} \leq \mathbf{c}_{slope} \leq c_{slope}^i+t_{slope}r_{slope}
    \end{dcases}
    \end{rcases},
\end{equation}
\noindent where $\mathbf{c}_{load}, \mathbf{c}_{slope}$ are vectors containing the history of load and slope contextual values, respectively. $c_{load}^i$ and $c_{slope}^i$ represent the contextual values for the current trial, and the variables $r_{load}(=40), r_{slope}(=0.5)$ denote the ranges of the load context and slope context. The threshold values are defined as $t_{load}=0.2$, $t_{slope}=0.1$.  

\begin{figure}[!bh]
\vspace*{-0.6em}
    \centering
    \includegraphics[clip, trim=1.0cm 3cm 1.0cm 2cm,width=0.9\linewidth]{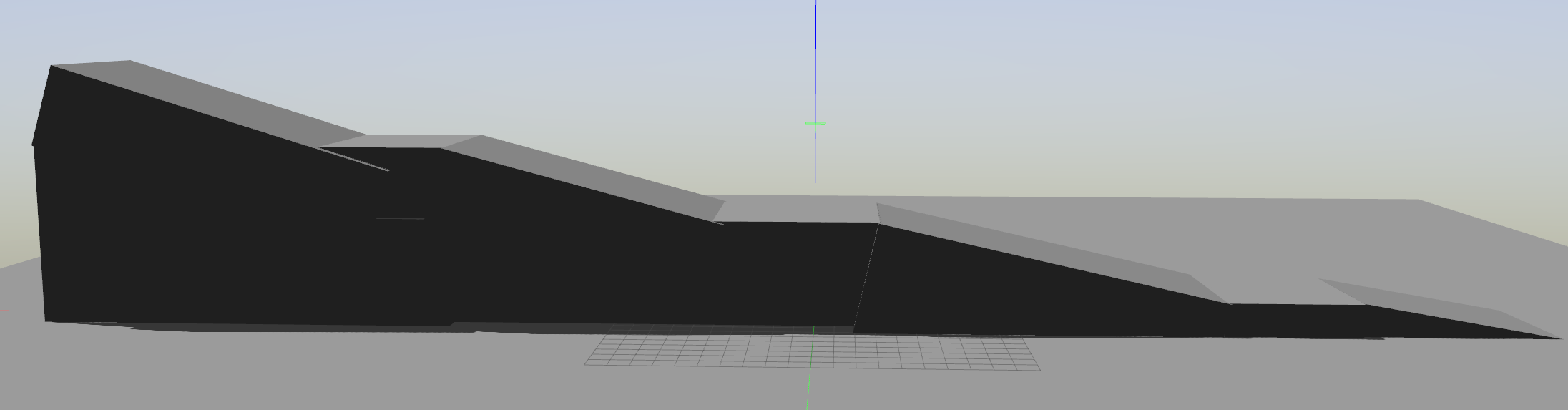} \\
    \vspace{-0.2em}
    \caption{Simulation settings for inference mode of slope adaptation in Gazebo.}
    \label{fig:slope_setting}
\vspace*{-2em}
\end{figure}

\subsection{Terrain Setup for Evaluating a Slope Adaptation Model}
As mentioned in Section \ref{sec:slope_load_adapt}, the model that had been optimized for flat terrain, as well as slopes of 10 degrees and 12.5 degrees, was further evaluated on a terrain that featured a continuous range of slopes from 10 degrees up to 17.5 degrees. The terrain setup for this evaluation is shown in Figure \ref{fig:slope_setting}.

\end{document}